\documentclass[sigconf]{acmart}
\def\BibTeX{{\rm B\kern-.05em{\sc i\kern-.025em b}\kern-.08emT\kern-.1667em\lower.7ex\hbox{E}\kern-.125emX}}
\usepackage{booktabs} 
\usepackage{caption}
\usepackage{subfigure}
\usepackage{graphicx}
\usepackage{subfloat}
\usepackage{multirow}
\usepackage{algorithm}
\usepackage{algorithmic}
\usepackage{tabularx}
\usepackage{array}
\usepackage{enumitem}
\usepackage{float}
\usepackage{soul}
\usepackage{xcolor}
\usepackage[normalem]{ulem}
\usepackage{url}

\usepackage{verbatim}

\newcolumntype{P}[1]{>{\centering\arraybackslash}p{#1}}
\newcolumntype{M}[1]{>{\centering\arraybackslash}m{#1}}

\newcommand*\eg{\textit{e.g.}}
\newcommand*\ie{\textit{i.e.}}

\setcopyright{acmcopyright} 
\begin{document}
\fancyhead{}
%
\title{Job2Vec: Job Title Benchmarking \\ with Collective Multi-View Representation Learning}

\author{Denghui Zhang$^1$, Junming Liu$^1$, Hengshu Zhu$^{2*}$, Yanchi Liu$^1$}
\author{Lichen Wang$^3$, Pengyang Wang$^4$, Hui Xiong$^{1,2}$}\authornote{Hui Xiong and Hengshu Zhu are corresponding authors. This work is supported by NSFC 91746301. The code is available at: https://github.com/zdh2292390/Job2Vec-Job-Title-Benchmarkingwith-Collective-Multi-View-Representation-Learning\vspace{-2mm}}


\affiliation{%
  \institution{$^1$Magagement Science and Information Technology Department, Rutgers University, USA\\
  $^2$Baidu Talent Intelligent Center, Baidu Inc, China \\
    $^3$Electrical and Computer Engineering Department, Northeastern University, USA\\
    $^4$Computer Science Department, University of Central Florida, USA\\}
}
\renewcommand{\shortauthors}{}

\copyrightyear{2019} 
\acmYear{2019} 
\acmConference[CIKM '19]{The 28th ACM International Conference on Information and Knowledge Management}{November 3--7, 2019}{Beijing, China}
\acmBooktitle{The 28th ACM International Conference on Information and Knowledge Management (CIKM '19), November 3--7, 2019, Beijing, China}
\acmPrice{15.00}
\acmDOI{10.1145/3357384.3357825}
\acmISBN{978-1-4503-6976-3/19/11}

\begin{abstract}
Job Title Benchmarking (JTB) aims at matching job titles with similar expertise levels across various companies. 
JTB could provide precise guidance and considerable convenience for both talent recruitment and job seekers for position and salary calibration/prediction. 
Traditional JTB approaches mainly rely on manual market surveys, which is expensive and labor intensive.
Recently, the rapid development of Online Professional graph has accumulated a large number of talent career records, which provides a promising trend for data-driven solutions.
However, it is still a challenging task since 
(1) the job title and job transition (job-hopping) data is messy which contains a lot of subjective and non-standard naming conventions for a same position (\eg, \textit{Programmer}, \textit{Software Development Engineer}, \textit{SDE}, \textit{Implementation Engineer}), 
(2) there is a large amount of missing title/transition information, 
and (3) one talent only seeks limited numbers of jobs which brings the incompleteness and randomness for modeling job transition patterns. 
To overcome these challenges, we aggregate all the records to construct a large-scale Job Title Benchmarking Graph (Job-Graph), where nodes denote job titles affiliated with specific companies and links denote the correlations between jobs.
We reformulate the JTB as the task of link prediction over the Job-Graph that matched job titles should have links. 
Along this line, we propose a collective multi-view representation learning method (Job2Vec) by examining the Job-Graph jointly in
(1) graph topology view (the structure of relationships among job titles),
(2) semantic view (semantic meaning of job descriptions), 
(3) job transition balance view (the numbers of bidirectional transitions between two similar-level jobs are close), and
(4) job transition duration view (the shorter the average duration of transitions is, the more similar the job titles are).
We fuse the multi-view representations in the encode-decode paradigm to obtain an unified optimal representations for the task of link prediction.
Finally, we conduct extensive experiments to validate the effectiveness of our proposed method. 
\end{abstract}


%
\vspace{-2mm}

%

%
\maketitle

\begin{figure}[t]
\includegraphics[width=0.40\textwidth]{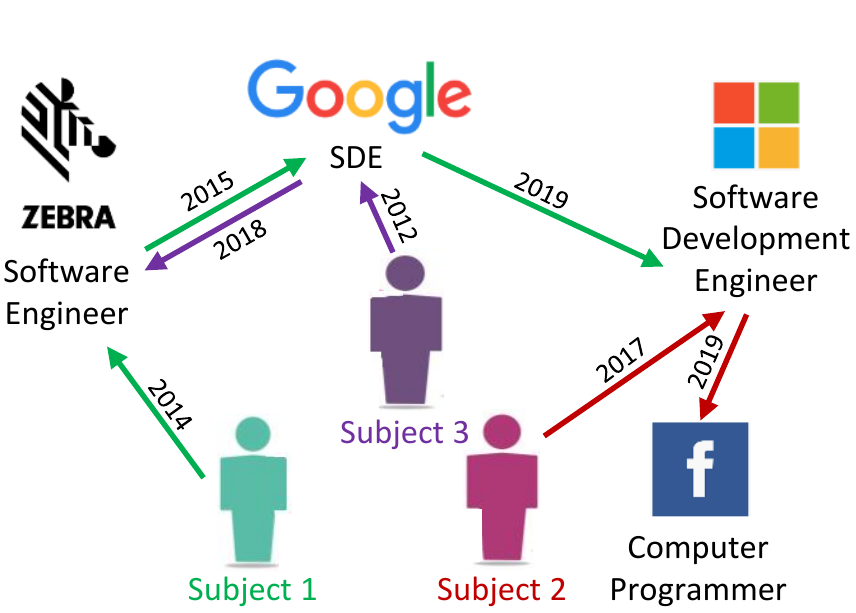}
\vspace{-3mm}
\caption{Job transitions of different subjects across companies and titles. Our approach aims to explore the multiple clues for high performance on job title benchmarking task.}
\label{fig:trajectory}
\vspace{-5mm}
\end{figure}

\section{Introduction}

Recent years have witnessed the increasing popularity of using data mining techniques for addressing human resource management (HRM) tasks (\eg, intelligent job-person fit and intelligent interview assessment~\cite{qin2018enhancing,shen2018joint}). However, few research efforts have been made on intelligent \textbf{J}ob \textbf{T}itle \textbf{B}enchmarking (JTB), which aims at matching job titles with similar expertise levels across various companies. For both job seekers and employers, JTB is important for talent recruitment and salary calibration. With appropriate JTB insights, employers can recruit relevant talent with the right title and salary. While for job seekers, JTB can provide guidance for their career development.
In this paper, we study the problem of JTB from the data mining persective.


Traditional JTB relies heavily on manual market surveys, which is expensive and labor intensive. Recently, the emergence of Online Professional graph (e.g., Linkedin) helps to accumulate a large number of career records, which provides an unparalleled opportunity for a data-driven solution.
However, JTB is still a challenging task due to the following three aspects.
First,  the job title and job transition (job-hopping) data is messy which contains a lot of subjective and non-standard naming conventions for the same position. 
For example, as shown in Figure~\ref{fig:trajectory}, Software Engineer, SDE, Software Development Engineer, and Computer Programmer are the same-level jobs across different IT companies.
Second, there is a large amount of missing title/transition information. Many users on Online Professional graph do not update their information in time. Too much missing information will hinder the applicability of data mining algorithms.
Third, in individual career, one talent only seeks limited numbers of jobs compared to the total set of job titles on the job market, which brings the incompleteness and randomness for modeling job transition patterns.

\begin{figure*}[h]
\includegraphics[width=1.0\textwidth]{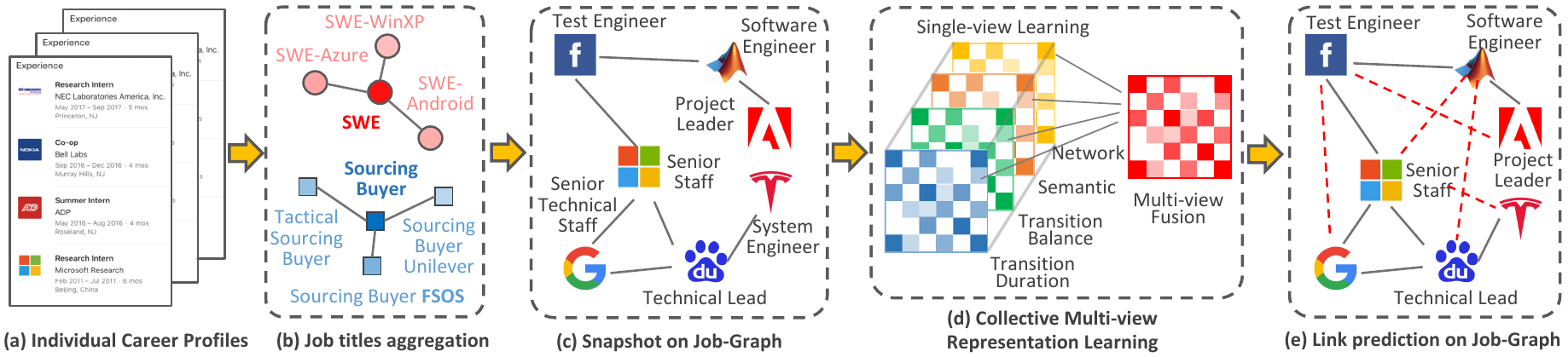}
\vspace{-7mm}
\caption{Overview of our job title benchmarking framework.}
\label{fig:framework}
\vspace{-3mm}
\end{figure*}

To tackle these challenges, we propose to construct a \textbf{J}ob \textbf{T}itle \textbf{B}enchmarking \textbf{G}raph (Job-Graph), where nodes denote job titles affiliated with the specific companies and links denote the numbers of transition between job titles. 
We hold the assumption that the bechmarked job title pairs should have strong correlations that there exists links between the job titles. 
Along this line, we reformulate JTB as the task of the link prediction over the Job-Graph.

Representation learning methods achieve outstanding performances for the link prediction task~\cite{zhang2017bl,wang2017predictive}.
However, due to the three unique properties of Job-Graph (the topology structure, rich semantic information of job titles, and job transition patterns),
existing representation learning methods are unable to model these properties at the same time.
Therefore, we propose a collective multi-view representation learning method to learn the representations of job titles for the task of link prediction.

Specifically, first, we model four views of representations:
(1) \textbf{Graph Structure View}, which refers to the topology structure of the Job-Graph that encodes the graph structure and neighborhood information;
(2) \textbf{Semantic View}, which refers to the semantic meaning of job titles;
(3) \textbf{Job Transition Balance View}, which is compliant with the observation that the numbers of bidirectional transitions between two similar-level jobs are close;
(4) \textbf{Job Transition Duration View}, which reveals the fact that the shorter the average duration of transitions is, the more similar the job titles are.
Then, to obtain an unified representation, we design a representation fusion process based on the encode-decode paradigm.
The multi-view representations are fed into the associated multi-layer perceptrons which are attached behind with a representation ensemble layer to work as an encoder.
The ensembled representation is dispatched to the corresponding decoder by a representation dispatching layer to reconstruct the multi-view representations.
The loss between the original and reconstructed multi-view representations will be minimized to guarantee the optimal unified representation.
Moreover, we train the multi-view representation learning procedure and the representation fusion procedure in an alternative way. 
The loss from four views and the representation fusion procedure will be minimized simultaneously to generate high quality representation for job titles for the task of link prediction.

\begin{table}[t]
\setlength{\textfloatsep}{0.3cm}
  \caption{An example of job transitions.}
  \label{tab:trajectory}
  \vspace{-3mm}
\scalebox{0.92}{  
  \begin{tabular}{|c|c|c|}
    \hline
    Job Title&Company&Duration\\
    \hline
    Production Engineer&Square Inc&2011/7-2016/10\\
    Senior Site Reliability Engineer & Google& 2010/10-2011/7\\
    Architect & Yahoo!& 2009/7-2010/6\\
  Systems Engineer & Yahoo! & 2006/6-2009/7\\
    Systems Engineer & IBM& 2006/2-2006/6\\
    \hline
\end{tabular}
}
\end{table}

In summary, in this paper, we propose a data-driven solution for the problem of JTB. 
Specifically, we first construct the Job Title Benchmarking Graph based on the job transition records. 
Then, we reformulate the problem of JTB as the task of link prediction.
We propose a collective multi-view representation learning method, by learning an unified job title representation from the graph structure view, semantic view, job transition balance view, and job transition duration view.
Finally, we conduct extensive experiment to evaluate our proposed method over the real-world dataset.
The promising results validate the effecitveness of our proposed method.

\section{Preliminaries}
In this section, we first briefly introduce the real-world dataset we collected for JTB task. 
Then, we introduce some essential definitions.
Followed the definitions, we propose our problem statement.
Finally, we present an overview of the propposed framework.

\subsection{Data Description}
In this study, we analyze real-world talent job title transition data, collected from a major commercial Online Professional Network. The data includes two main categories, IT-related and Finance-related job titles. Table~\ref{tab:stat} shows Job-Graphs constructed from these two datasets are very sparse.

Table~\ref{tab:trajectory} presents an example of  job transition records from an individual talent. Each line consists of a job title, company name and the duration holding this position.

\begin{table}[t]
  \caption{Statistics of Job-Graphs in IT and Finance.}
  \vspace{-3mm}
  \label{tab:stat}
\scalebox{0.92}{   
  \begin{tabular}{|c|c|c|}
    \hline
    Dataset &IT & Finance\\
    \hline
    \# Edges &39927& 77118\\
    \# Nodes & 44030 &89851\\
    Average out degree & 0.91&0.86\\
    \hline
\end{tabular}
}
 \vspace{-5mm}
\end{table}

\subsection{Definition}
Here, we introduce some essential definitions, which will be used throughout this paper.

\begin{definition}{\textbf{Job Title Bechmarking (JTB)}}
JTB is a process that matches job titles with similar expertise levels across various companies. Formally, given two job title-company pair, $(Title_i, $ $Company_i)$ and $(Title_j, Company_j)$, the objective is determine whether the given paired job titles are on the same level. JTB could provide precise guidance and considerable convenience for both talent recruitment and job seekers for position and salary calibration/prediction.
\end{definition}

\begin{definition}{\textbf{Job Title Transition Graph (Job-Graph).}}
Job-Graph is defined as a directed graph $G=(V, E)$, where each node $v_i \in |V|$ represents a job title affiliated a company(i.e., $v_i=(Title_i, Company_i)$), and each link $e_{ij} \in |E|$ between two nodes $v_i$ and $v_j$ indicates that there exist job transitions from the $(Title_i, $ $Company_i)$ to $(Title_j, Company_j)$, and the weight of edge $e_{ij}$ represents the number of transitions observed from $(Title_i, $ $Company_i)$ to $(Title_j, Company_j)$ 


\end{definition}

\subsection{Problem Statement}

In this paper, we study the problem of Job Title Benchmarking (JTB). 
We first construct job title transition graph to depict the job transition patterns.
We formulate the JTB as the the task of link prediction over the Job-graph, based on the assumption that similar-level job titles should have strong correlations to enable a link between them.
To enable the link prediction task, we push forward the problem formulation to the representation learning over the Job-Graph to learn unified and optimal representations for job titles.

Formally, given the Job-Graph $G=(V, E)$, we aim to find a mapping function $f: \mathbf{v} \rightarrow \mathbf{z}$ that takes node (job title) $\mathbf{v}$ as the input, and outputs the vectorized representation $\mathbf{z}$ of the job titles, while preserving properties of the Job-Graph and job transition patterns. The generated node (job title) representation $\mathbf{v}$ is then utilized to solve the problem of link prediction.

\theoremstyle{definition}

\vspace{-0.3cm}
\subsection{Framework Overview}
Figure~\ref{fig:framework} shows an overview of our proposed framework that includes the following essential tasks: (i) constructing the job title transition graph; (ii) developing a collective multi-view representation learning method for learning job title representations; (iii) applying learned job title representations for link prediction on Job-Graph. In the first task, given job transition records of talents, we construct a job title transition graph. In the second task, a collective multi-view representation learning method is developed for jointly modeling the graph structure, semantic meaning of job titles and job transition patterns. In the last task, we apply our proposed method to learn job title representations for link prediction on Job-Graph to benchmark job titles.

\section{Job-Graph Construction and Refinement}

In this section, we show how to construct the Job-Graph. Intuitively, the Job-Graph can be directly constructed from the raw job tranistion record data. However, messy, noisy and non-standard name convention of job titles makes the Job-Graph extremely sparse and redundant, which hinders the further analysis. Therefore, we refine the Job-Graph into an applicable fashion in the following steps: (1) Extract job transitions from the raw career records; (2) Map and aggregate all the transitions into the Job-Graph, where each node represents a job title, each edge represents the number of transitions between the nodes.

\subsection{Job Transition Extraction}
As shown in Table~\ref{tab:trajectory}, each transition consists of a source job and destination job, i.e., $(job_{src}, job_{des})$. We first set all the job titles existed in the raw data as nodes. Then we sum the transition frequencies as the link from $job_{src}$ to $job_{des}$. The constructed graph is set as the base graph for the further refinement.

\subsection{Job Title Aggregation}
\begin{figure}[t]
\setlength{\belowcaptionskip}{-7pt}
\includegraphics[width=0.48\textwidth]{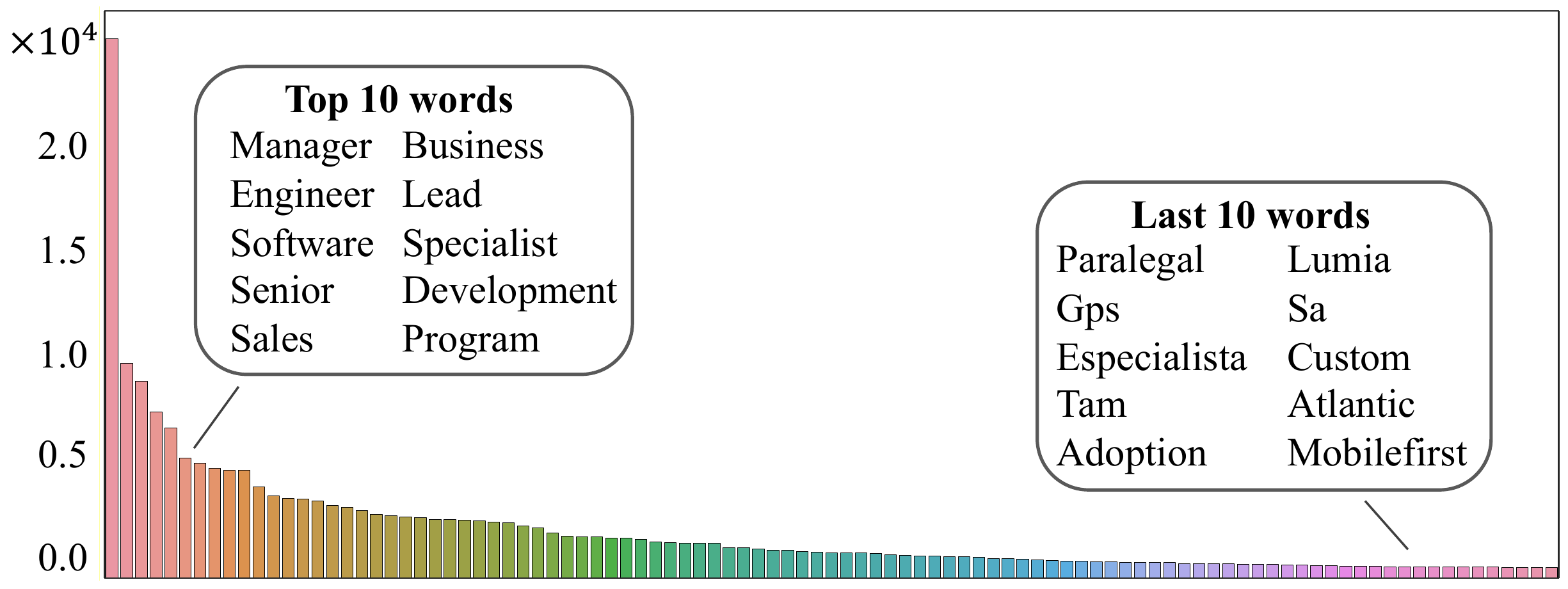}
\vspace{-0.6cm}
\caption{Words frequency distribution.}
\label{fig:wordfreq}
\vspace{-0.3cm}
\end{figure}

Generally speaking, job titles usually consist of three parts:
\begin{enumerate}
    \item \textbf{Title level}, such as \textit{Senior}, \textit{Principle} and \textit{Director}.
    \item \textbf{Title core function}, such as Software Engineer, Product Manager.
    \item \textbf{Unique additional information}, such as  Software Engineer  \textit{in Bid Data}, Sales Rep  \textit{on Small and Medium businesses}.
\end{enumerate}
After we study the word frequencies of job titles, there is an interesting observation that the word frequency distribution  subjects to {\it power law distribution}, as shown in Figure \ref{fig:wordfreq}. It also can be observed that noise words or those additional user-related words usually appear in the long tail. We also show the top 10 frequent words and bottom 10 frequent words. It is obvious that the most frequent words like \textit{manager} and \textit{engineer} usually describe the core function of job titles, while the less frequent words looks more like user's unique information. With this observation, we aggregate job titles by filtering out low frequency words in job titles and thus get a normalized and denser Job-Graph. Specifically, in this work, we filtered out the words that have a frequency lower than 30.

Table \ref{tab:filter} shows three real examples of aggregated job titles by filtering out low frequency words. Words in bold indicate the low-frequency words. For example, \textit{"Tactical Sourcing Buyer (Unilever)"} and \textit{"Sourcing Buyer, MARCOM \& FSOS"} are originally thought to be different, after filtering low-frequency words, they are aggregated to the same title \textit{"Sourcing Buyer"}. To be noted, the reason we use filtering low-frequency words instead of using cluster algorithm to cluster the job titles is that there are no standard target classes for job titles, and it is hard to decide the number of clusters for job titles if cluster algorithm is applied.

With this aggregated Job-Graph, we can obtain some concise job title matching insights like: A Senior Software Engineer of LinkedIn can match a Software Engineer of Google since most Senior Software Engineers of LinkedIn obtained the title of  Software Engineer when they just made a career transition to Google. However, the sparsity issue of Job-Graph still limits the performance of traditional representation learning method. Therefore, in next section, we introduce our collective multi-view representation learning method, Job2Vec, and show how to perform link prediction to enrich Job-Graph based on the proposed method.

\begin{table}[t]
\caption{Examples of aggregating job titles.}
\vspace{-2mm}
\label{tab:filter}
\setlength\tabcolsep{0.5pt}
\scalebox{0.92}{  
\begin{tabularx}{\linewidth}{ |P{5.0cm}|c|} 
\hline
Original Job Titles & Aggregated Job Title  \\
\hline
\textbf{Tactical} Sourcing Buyer \textbf{(Unilever)} 
& \multirow{2}{*}{Sourcing Buyer} \\[1pt]
 \cline{1-1} 
Sourcing Buyer, \textbf{MARCOM \& FSOS} & \\[1pt]
\hline
 Software Design Engineer-& \\
\textbf{(Azure)}&\multirow{4}{*}{Software Design Engineer} \\
\cline{1-1}
Software Design Engineer-& \\
\textbf{WindowsXP}& \\ 
\cline{1-1}
Software Design Engineer-&\\ 
\textbf{(Contracting) Encarta}&\\ 
\hline
 \textbf{Cyber} Security Architect&\multirow{2}{*}{Security Architect} \\[1pt] 
 \cline{1-1}
Security Architect&\\
\hline
\end{tabularx}
}
\vspace{-3mm}
\end{table}

\section{Job2Vec: Collective Multi-view Representation Learning for Job Titles}
In this section, we introduce our collective multi-view representation learning methods fore job titles.

\subsection{Model Intuition}
We learn representations of job titles with the following intuitions.

\textbf{Intuition 1: Topology Structure Preservation.} Job-Graph is built to depict job transitions on the job market. 
The topology structure of Job-Graph reveal the connectivity and neighbor information of job titles which can help to describe the latent structures among job titles.
We should preserve the topology structure of job-graph in the representation learning.

\textbf{Intuition  2: Semantics Preservation.} Job titles contain rich semantic descriptions which can further enhance the quality of job title representation. Therefore, we should preserve the semantic meanings of job titles.

\textbf{Intuition 3: Job Transition Patterns Preservation.} Job transitions have unique latent patterns. Transitions among different job title pairs are also different. Consequently, we should preserve the job transition patterns.

Therefore, we model the job title representations in multi views. Specifically, we introduce {\it graph topology view} for Intuition 1, {\it semantic view} for Intuition 2, {\it job transition balance view} and {\it job transition duration view} for Intuition 3. In addition, we propose a collective method to fuse the multi-view representations into an unified representation. We introduce the details as follows.

\subsection{Graph Topology View}
Job graph structure explicitly illustrates the similarity and correlations between different titles. It is the most crucial and effective information which provides comprehensive and accurate title connections. However, the topology information is hidden in the graph structure. Motivated by the success of graph representation learning methods in link prediction on social graph and knowledge graph~\cite{leskovec2010signed,zhang2017efficient,li2018link,wang2016structural,chang2015heterogeneous}, the first view we use in Job2vec is the \textbf{Graph Topology} view, which can encode the graph structure and neighborhood information into the representations. In Graph Topology view, we aim to learn a low-dimensional representation for each job title which can keep the neighborhood structure information, \ie, job titles that share similar neighbors in Job-Graph should be close to each other in the graph view representation space.

To achieve this, we first assign each job title $v_i$ into two representation vectors, "self representation", $e_i$ $\in \mathbb{R}^{N_g}$, and "neighbor representation", $e'_i$ $\in \mathbb{R}^{N_g}$, where $N_g$ is dimension of the representation vector of Graph Topology View. Both $e_i$ and $e'_i$ are randomly initialized. We utilize its "self representation" while "neighbor representation" will be used if $v_i$ is just the neighbor of the one we focus on. Then, in order to enforce embeddings to be close to each other if they share similar graph neighbors, we define the loss function $O_{N}$ as followed:
\begin{equation}
O_{N}=-\sum_{(i,j)\in E}w_{ij}log(p(v_j | v_i)),
\end{equation}
where $w_{ij}$ is the weight between $v_i$ and $v_j$, $E$ is the set of all edges in Job-Graph, to incorporate high-order proximity, we extend $E$ by adding edges of k-steps paths into the set. When $k=1$, $O_{N}$ is the same as the second-order proximity in LINE. $p(v_j | v_i)$ is the probability of $v_j$ occured as neighbor given $v_i$, defined as a softmax function followed:
\begin{equation}
p(v_j|v_i)=\frac{\exp(\vec{e}_j'^{T}\cdot \vec{e}_i)}{\sum_{k=1}^{|V|}\exp(\vec{e}_k'^{T}\cdot, \vec{e}_i)},
\end{equation}
where $v_i$ is the current job title we focus on, $v_j$ is the neighbors of $v_i$, $e_i$ is the "self representation" of $v_i$, $e'_j$ is the "neighbor representation" of $v_j$, $V$ is the set of all job titles, \ie, all nodes in Job-Graph, $|V|$ is the cardinal number of $V$. Minimizing $O_{N}$ equals to maximizing the conditional probability of $v_j$ given $v_i$. Since the conditional probability $p(v_j | v_i)$ is parameterized by $e'_j$ and $e_i$, as a result, the "self representation" of those job titles that share similar neighbors will be similar, \ie, close in the graph view space. To be noted, in testing stage, we utilizes the "self representation" of each job title to calculate the similarity score.

\subsection{Semantic View}

Normally, each job title is consisted of several key words which describe the basic function and duty of this job (\eg, Project Manager and Computer Engineer). Therefore, the semantic information contained in these keywords is crucial to be explored in the representation learning process for two reasons: (1) Talents tend to make transition between functionally similar jobs, and semantic information guides the model to learn a better representation to tackle the complex job transition pattern. (2) The shared key words in job titles could further connect them even though the Job Transition Graph is extremely sparse, thus can alleviate the sparsity issue and improve the prediction capability of the learned representation.

We consider this view as the semantic view of Job Transition Graph. In semantic view, we aim to learn a low-dimensional representation of each job title which can keep the semantic information, \ie, job titles that have similar key words should be close to each other in the semantic view representation space. To achieve this, we first assign each job title $v_i$ a vector $s_i$ $\in \mathbb{R}^{N_s}$, and each word $w_j$ in the Job-Graph vocabulary a vector $s'_j$ $\in \mathbb{R}^{N_s}$ which are randomly initialized. $N_s$ is dimension of the representation vector of Semantic View. Then, we enforce $s_i$ to be close to each other if they share similar words, based on the loss function $O_{S}$ defined as followed:
\begin{equation}
O_{S}=-\sum_{w_j\in v_i}f_{ij}log(p(w_j | v_i)),
\end{equation}
where $f_{ij}$ is the frequency of the word $w_j$ occurred in $v_i$, $v_i$ is a job title, $w_j$ is the words in $v_i$, $p(w_j | v_i)$ is the probability of $w_j$ occurred in  $v_i$, defined as a softmax function:
\begin{equation}
p(w_j|v_i)=\frac{\exp(\vec{s}_j'^{T}\cdot \vec{s}_i)}{\sum_{k=1}^{|W|}\exp(\vec{s}_k'^{T}\cdot \vec{s}_i)},
\end{equation}
where $W$ is the vocabulary set of Job-Graph, $s_i$ is the semantic representation of job title $v_i$, $s'_j$ is the semantic representation of word $w_j$. Minimizing $O_{S}$ equals to maximizing the conditional probability of $w_j$ given $v_i$, as a result, job titles $v_i$ and $v_j$ will have similar representations $s_i$ and $s_j$ if they are similar.


\subsection{Job Transition Balance View}
The numbers of bidirectional transitions between two similar-level jobs are close. For example, software engineer in Apple is on the same level as the SDE in Facebook. 
The transition number from software engineer (Apple) to the SDE (Facebook), and the transition number from the SDE (Facebook) to software engineer (Apple) should be close.
However, for two different-level jobs, like junior software engineer and senior software engineer, the transition usually is in one direction, from the junior to the senior one. In other words, the transition number for two-directions will be very different.
To this end, we further consider Job Transition Balance as an important factors for JTB, which effectively indicates the matches of each pair of titles. The intuition of Transition Balance is that if comparable amounts of talent transitions can be found in both direction between two job titles, then these two job titles are highly likely to be in the same level. To model Transition Balance, we first assign each job title $v_i$ a vector $b_i$ $\in \mathbb{R}^{N_b}$ which is randomly initialized. $N_b$ is the dimension of the representation vector of Transition Balance View. Then given two job titles $v_i$ and $v_j$, we define the Transition Balance (TB) between them as:
\begin{equation}
TB(v_i,v_j) = \exp(-\frac{|w_{ij}-w_{ji}|}{w_{ij}w_{ji}}),
\end{equation}
where $w_{ij}$ is the weight of the edge from $v_i$ to $v_j$. Then, based on the loss function $O_{B}$ we enforce $b_i$ to be similar to each other if they have balanced transitions between each other. $O_{B}$ is defined as followed:
\begin{equation}
O_{B}=-\sum_{(i,j)\in v_i}TB(v_i,v_j)log(p(v_i,v_j)).
\end{equation}
where $p(v_i,v_j)$ is the joint probability of $v_i$ and $v_j$ defined as:
\begin{equation}
p(v_i,v_j)=\frac{1}{1+\exp(-\vec{b}_i^{T}\cdot \vec{b}_j)}.
\end{equation}

Minimizing $O_{B}$ will "drag" the representation vectors of those "balanced" job title pair to be close in the representation space.

\begin{figure}[t]
\setlength{\belowcaptionskip}{-7pt}
\includegraphics[width=0.47\textwidth,height=0.32\textwidth]{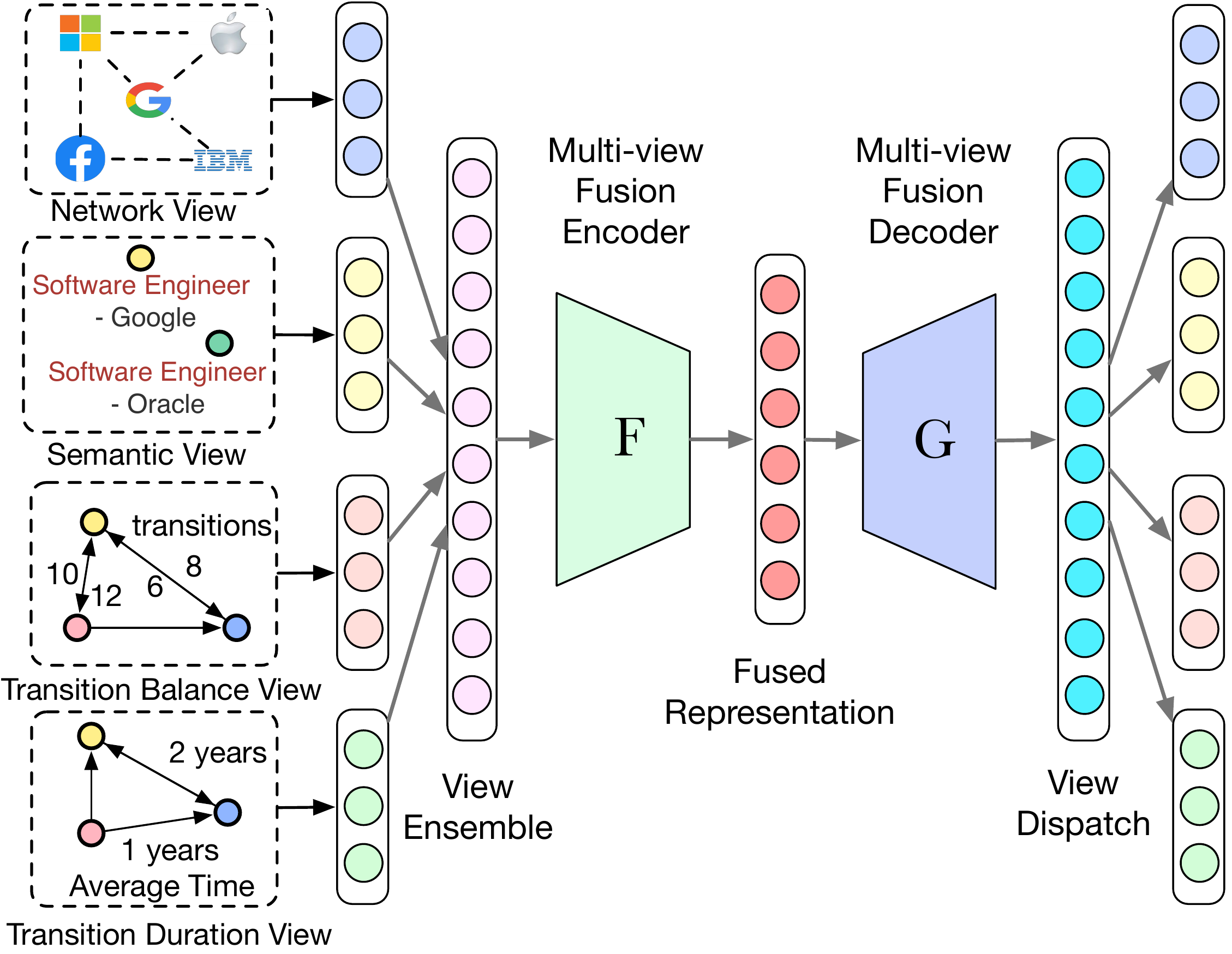}
\vspace{-3mm}
\caption{Collective Multi-View Representation Learning}
\label{fig:fusion}
\end{figure}
\subsection{Job Transition Duration View}
Most people require a relatively long time (\eg, one or several years) to get a promotion. In contrast, if a person can change his/her jobs quickly and frequently, then there are high possibilities that these jobs are similar titles requires similar expertise and working experience. Make a long story short, the shorter the average duration of transitions is, the more similar the job titles are. To this end, we define Job Transition Duration which is the average duration time between two job titles. To include Transition Duration property into our model, we first assign each job title $v_i$ a vector $d_i$ $\in \mathbb{R}^{N_d}$ which are randomly initialized. $N_d$ is dimension of the representation vector of Transition Duration View. Given two job titles $v_i$ and $v_j$, we define the Transition Duration (TD) between them as:
\begin{equation}
TD(v_i,v_j) = \exp(-t_{ij}),
\end{equation}
where $t_{ij}$ is the average duration time from $v_i$ to $v_j$. Then we designed a loss function $O_{D}$, which enforces $d_i$ and $d_j$ to be closer to each other if the average transition time between them is short, \ie, it is easy to transit from $v_i$ to $v_j$. $O_{D}$ is defined as:
\begin{equation}
O_{D}=-\sum_{(i,j)\in v_i}TD(v_i,v_j) log(p(v_i,v_j)),
\end{equation}
where $p(v_i,v_j)$ is the joint probability of $v_i$ and $v_j$ defined as:
\begin{equation}
p(v_i,v_j)=\frac{1}{1+\exp(-\vec{d}_i^{\top}\cdot \vec{d}_j)}.
\end{equation}

Minimizing $O_{D}$ will "drag" the representation vectors of those "easily transited" job title pair to be close in the representation space, which hopefully further improve the learning performance.


\subsection{Multi-view Representation Fusion}
Each views provides a comprehensive and unique aspect across different titles, and there are more informative and sophisticated correlation knowledge residing across different views. While, naively combining all the views together cannot efficiently utilize the information, or even hurt the performance due to the dramatic different scales and formats of the views.

To this end, we propose a Collective Multi-View Auto-Encoder (CMVAE) framework to compress the multiple representations into a single denser representation. As shown in Figure \ref{fig:fusion}, the four different representations obtained by learning from the above mentioned objectives are feed into the CMVAE. To avoid losing information from all the representations, we directly concatenate the representations and feed them into the Fusion Encoder $\mathcal{F}(\cdot)$. $\mathcal{F}(\cdot)$ consists of two fully-connected layers and outputs a single and denser representation. Then this intermediate representation is feed to the Fusion Decoder $\mathcal{G}(\cdot)$ which is also a two fully-connected layer neural network and outputs the restored representation. The objective function of CMVAE is shown below:
\begin{equation}
L = \frac{1}{N}\sum_{i=1}^N\| X_i-\mathcal{G}(\mathcal{F}(X_i)) \|_2^2,
\end{equation}
where $N$ is the number of the training samples. $X_i = [e_i; s_i; b_i; d_i]$ is the ensembled multi-view representation for a job title $v_i$,  $\mathcal{G}(\mathcal{F}(X_i))$ is the restored representation. Minimizing the difference between the raw representations $X_i$ and restored representations $\mathcal{G}(\mathcal{F}(X_i))$ will enforce the model to learn a denser and unified representation $\mathcal{F}(X_i)$. CMVAE hopefully captures the distinctive aspects from different views and further reveals the latent correlations across the views. We jointly optimize CMVAE associate with the other individual-view representation graphs. This jointly training strategy could let each graph assistant other graphs and further enhance the learning performance. Finally, we use $\mathcal{F}(X_i)$ as the fused multi-view representation for subsequent link prediction task.

\section{Experiment Results}
This section details our empirical evaluation of the proposed method on real-world data.

\subsection{Experimental Data}
Table \ref{tab:data} presents the statistics of our data sets from Information Technology industry and Finance industry. Here we provide more details about our real-world data as follows:

\noindent\textbf{IT Data.} 
To construct IT data, we randomly sampled one million user career records who have been working at several well-known IT companies in US. For ease of analysis, we chose 15 most famous and leading companies in IT and only keep transition records of these companies. The 15 companies include \textit{Facebook, Google, Amazon, Microsoft, Apple, IBM, LinkedIn, Cisco, Oracle, Airbnb, Uber, Yahoo, Nokia, Apple, Intel, HP.}
Then we used the methods mentioned and Section 3 to construct the Job-Graph and aggregated it. Finally we got the IT Job-Graph shown in Table \ref{tab:data}. The time span of the data is from 06/18/1998 to 12/01/2018. 

\noindent\textbf{Finance Data.} For finance data, we randomly sampled one million records according to the rule: whose records contains finance related keywords such as \textit{Finance, Asset Manager, Financial Research Analyst, Investment Banking Analyst, Equity Research Analyst, Trust Officer, Commercial Banker, etc.} Then again we used the methods mentioned in Section 3 to construct the Job-Graph and aggregated it. Finally we got the Finance Job-Graph shown in Table \ref{tab:data}. The time span of the data is from 03/20/2004 to 12/01/2018. 

\begin{table}
\caption{Statistic Details of the Dataset}
\vspace{-4mm}
\centering
\label{tab:data}
\begin{tabular}{|M{2cm}|P{1.2cm}|P{1.2cm}|P{1.2cm}|P{1.2cm}|}
\hline
& \multicolumn{2}{c|}{\textbf{IT}}  & \multicolumn{2}{c|}{\textbf{Finance}} \\

\hline
& train & test & train & test   \\

\hline
\# Nodes & 44,030 & 2,838 &89,851  &5,512 \\
\hline
\# Edges & 38,133 & 1,794 &72,140 &4,978 \\

\hline
\# Transitions &\multicolumn{2}{P{2.4cm}|}{45,095} & \multicolumn{2}{P{2.4cm}|}{92,085} \\
\hline
\# Job titles &\multicolumn{2}{P{2.4cm}|}{44,030} & \multicolumn{2}{P{2.4cm}|}{89,851} \\
\hline
\# Companies & \multicolumn{2}{P{2.4cm}|}{15} & \multicolumn{2}{P{2.4cm}|}{7,669}\\
\hline
\# Time span & \multicolumn{2}{P{2.4cm}|}{1998/06-2018/12} & \multicolumn{2}{P{2.4cm}|}{2004/03-2018/12} \\
\hline
Job title examples & \multicolumn{2}{P{2.4cm}|}{Software Engineer Google} & \multicolumn{2}{P{2.4cm}|}{Asset Manager Goldman Sachs} \\

\hline
\end{tabular}
\end{table}

\begin{table*}[t]
\caption{Link Prediction Performance Comparison.}
\vspace{-4mm}
\centering
\label{tab:linkprediction}
\scalebox{0.92}{
\begin{tabular}{ |c|c|c|c|c|c|c|c|c|c|c| }

\hline
& \multicolumn{5}{c|}{IT Dataset}  & \multicolumn{5}{c|}{Finance Dataset} \\

\hline
& $MRR$ & $MP@5$ &  $MP@10$ &$MP@15$ &  $MP@20$ & $MRR$ & $MP@5$ &  $MP@10$ &$MP@15$ &  $MP@20$  \\

\hline
DeepWalk &0.0688  & 0.0858 & 0.1070 &0.1198  &0.1293  &0.1044 &0.1164  &0.1444 &0.1600 &0.1675  \\
\hline
Node2Vec & 0.0645 & 0.0785 &0.0925  &0.1042  &0.1153  &0.0979 &0.1065  &0.1235 &0.1335 &0.1460  \\
\hline
Line(1st order) & 0.0644 & 0.0752 &0.0947  &0.1081  &0.1198  &0.0983 &0.1071  &0.1245  &0.1341 &0.1428 \\
\hline
Line(1st+2nd order) & 0.0651 & 0.0791 &0.0958  &0.1064  &0.1125  &0.0943 &0.0994  &0.1150 &0.1274 &0.1381  \\
\hline
Word2Vec &0.1295  &0.2135  &0.2792  &0.3194  &0.3334  &0.1110  &0.1239  &0.1452  &0.1643  &0.1775 \\

\hline
Job2Vec & \textbf{0.1734} & \textbf{0.2391} &\textbf{0.2976}  &\textbf{0.3288}  &\textbf{0.3511}  &\textbf{0.1311} &\textbf{0.1378}  &\textbf{0.1635}  &\textbf{0.1815} & \textbf{0.1978}\\

\hline
\end{tabular}
}
\vspace{-2mm}
\end{table*}

\subsection{ Baselines \& Evaluation Metrics}
We compare our proposed method with the following representative baselines of representation learning:

\noindent\textbf{Deepwalk~\cite{perozzi2014deepwalk}:} DeepWalk adopted a truncated random walk on a graph to generate a set of walk sequences and train Skip-Gram on these sequences. It only considers graph topology.

\noindent\textbf{Node2Vec~\cite{grover2016node2vec}:} Node2Vec generalizes DeepWalk by defining a more flexible notion of a node's graph neighborhood. It only considers graph topology.

\noindent\textbf{LINE(1st order)~\cite{tang2015line}:} In LINE, first-order and second-order proximity are modeled by the joint probability distribution between two nodes and the similarity between their neighborhood respectively. LINE(1st order) only keeps first-order proximity. It only considers graph topology.

\noindent\textbf{LINE(1st+2nd order):} This is the full model of LINE. It keeps both first-order and second-order proximity. It only considers graph topology.

\noindent\textbf{Word2Vec\cite{mikolov2013distributed}:} Word2Vec only applies semantic view. Specifically, we treat each job title as a sentence, then train Word2Vec on all the job titles on Job-Graph. Finally we get the embedding vector of a job title by averaging the vectors of the words in it.

\noindent\textbf{Job2Vec:} The model proposed in this paper which considers four crucial aspects (graph topology, semantic, transition balance, transition duration) of the Job-Graph.


\noindent\textbf{Metrics:}  
We use two metrics, namely, $MRR$ and $MP@K$, to evaluate the link prediction performance.
\begin{itemize}[leftmargin=*]
\item For each test $i$, the correct answer job title is identified at position rank$[i]$ for closest job titles.
The \emph{Mean Reciprocal Rank (MRR)} is
\begin{equation}
\label{MRR}
MRR=\frac{1}{N}\sum_{i=1}^N \frac{1}{\text{rank}[i]}.
\end{equation}
Higher MRR means that correct answers appear more closely with the  query job title.

\item Additionally,  for test $i$ consisting of a query job title and target job title pair, consider the closest $K$ job titles to the query embedding. If the correct target job title to the query job title is among these $K$ titles, then the \emph{Precision@K} for test $i$ (denoted P@K[$i$]) is 1; else, it is 0.
Then the 
\emph{Mean Precision@K} is defined as
\begin{equation}
\label{precision}
MP@K = \frac{1}{N}\sum_{i=1}^N (P@K[i]).
\end{equation}

Higher precision indicates a better ability to acquire correct answers using close embeddings.
\end{itemize}

\subsection{Link Prediction Performance Comparison}

We performed link prediction on both IT and Finance datasets, i.e., predicting missing links on Job-Graph. Since edges with a larger weight in Job-Graph indicates a better match between job titles, we kept edges that have weight larger than a threshold (here is 5) for training and also try to predict links that have weight larger than 5 (predicted links if have weights lower than 5 will be considered as wrong). Then we randomly split the Job-Graph links into 10 equal parts, 8 of them as training set, 1 as validation set and 1 as testing set, no data in validation set and testing set can be used for training the embeddings. To avoid "cold start", we only kept job titles that have occurred in training data. We trained all the baseline models and our Job2Vec on the training set to get the embeddings, to avoid overfitting we tuned parameters on the validation set, finally we predict links on the testing set using the learned embeddings. Given a job title $job_i$, to predict which job may have links with it, we calculated the cosine similarity score between the embeddings of $job_i$ and the rest of other jobs, then ranked them based on the similarity score. Higher ranked jobs have a higher probability to be matched with $job_i$.

We obtained the best hyper parameters of our model on the validation set. The dimensions of the four views' representations are $N_g, N_s, N_b, N_d =128$. In the Collective Multi-View Auto-Encoder (CMVAE), for the fusion encoder, we use a two-layer fully-connected network (512*512*248) with LeakyReLU activation (negative-slope=0.7) in the first layer and Tanh activation before output. For the fusion decoder, we use a two-layer fully-connected network (248*512*512) with LeakyReLU activation (negative-slope=0.7) in the first layer and no activation in the second layer.

From the results on the IT dataset, as shown in Table \ref{tab:linkprediction}, we can tell that graph embedding models that only keep graph topology information get similar but poor performances, where DeepWalk achieves the best $MRR$ and $MP@K$. Word2Vec(averaging word vectors in job title) achieves nearly 100\% improvements on $MRR$ and on $MP@K$ compared with DeepWalk, thus proves that semantic view is tremendously helpful for link prediction on Job-Graph. Job2Vec achieves further improvements on $MRR$ compared with Word2Vec, which shows the effectiveness of preserving more views than only semantic view. Job2Vec achieves the best performance on $MRR$ and $MP@5$, $MP@10$, $MP@15$, $MP@20$ among all the models. Specifically, it improves 200\% over DeepWalk and 50\% over Word2Vec. This confirms the superiority of the multi-view representation and the effectiveness of the encode-decode paradigm for fusing multiple views. From the results on the Finance dataset, similar conclusions can be drawn.


\subsection{Robustness Analysis}



\begin{figure}
\subfigure[DeepWalk]{\label{fig:deepwalk-robust}\includegraphics[width=0.49\linewidth]{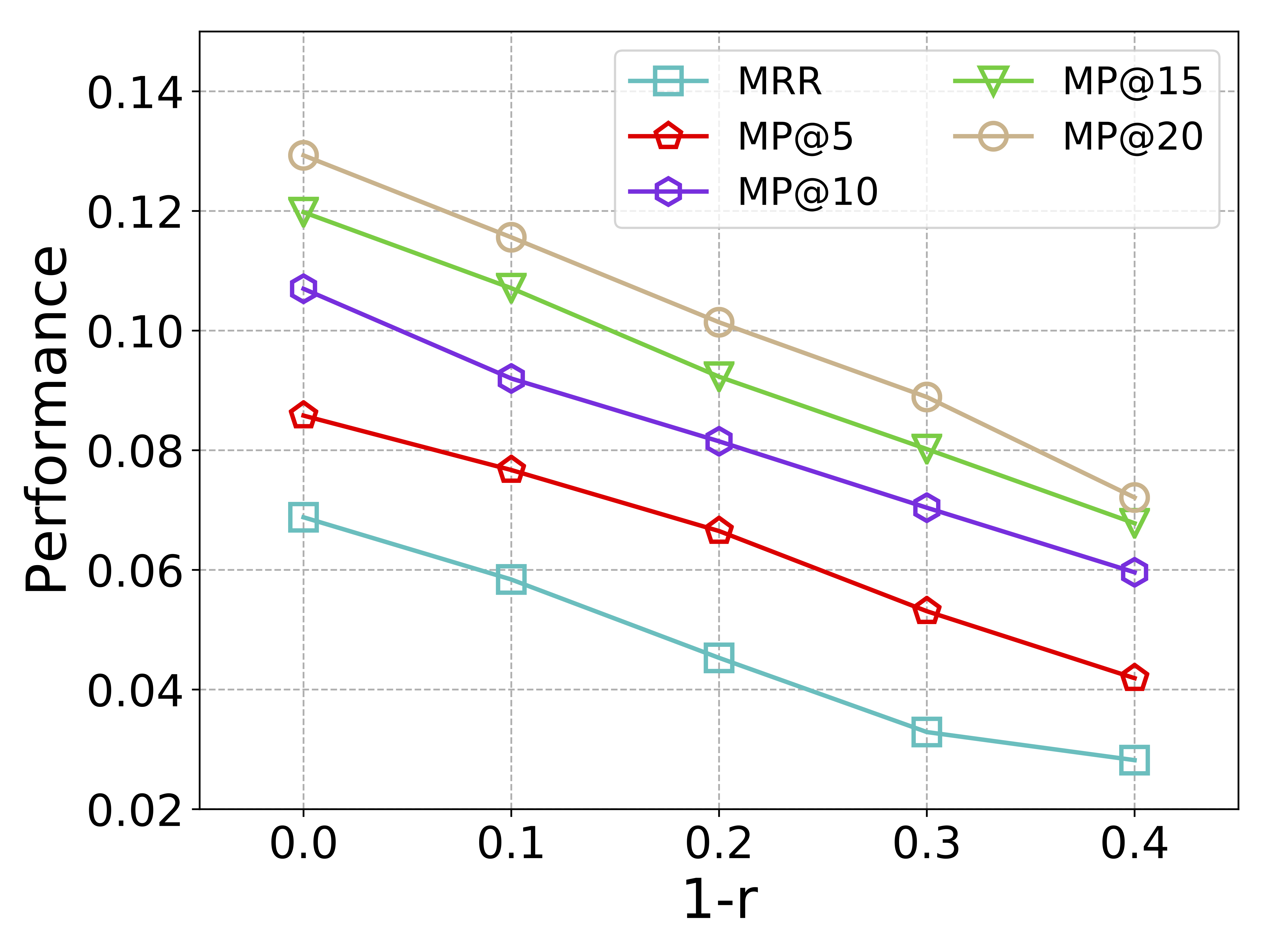}}
\subfigure[Job2Vec]{\label{fig:Job2Vec-robust}\includegraphics[width=0.49\linewidth]{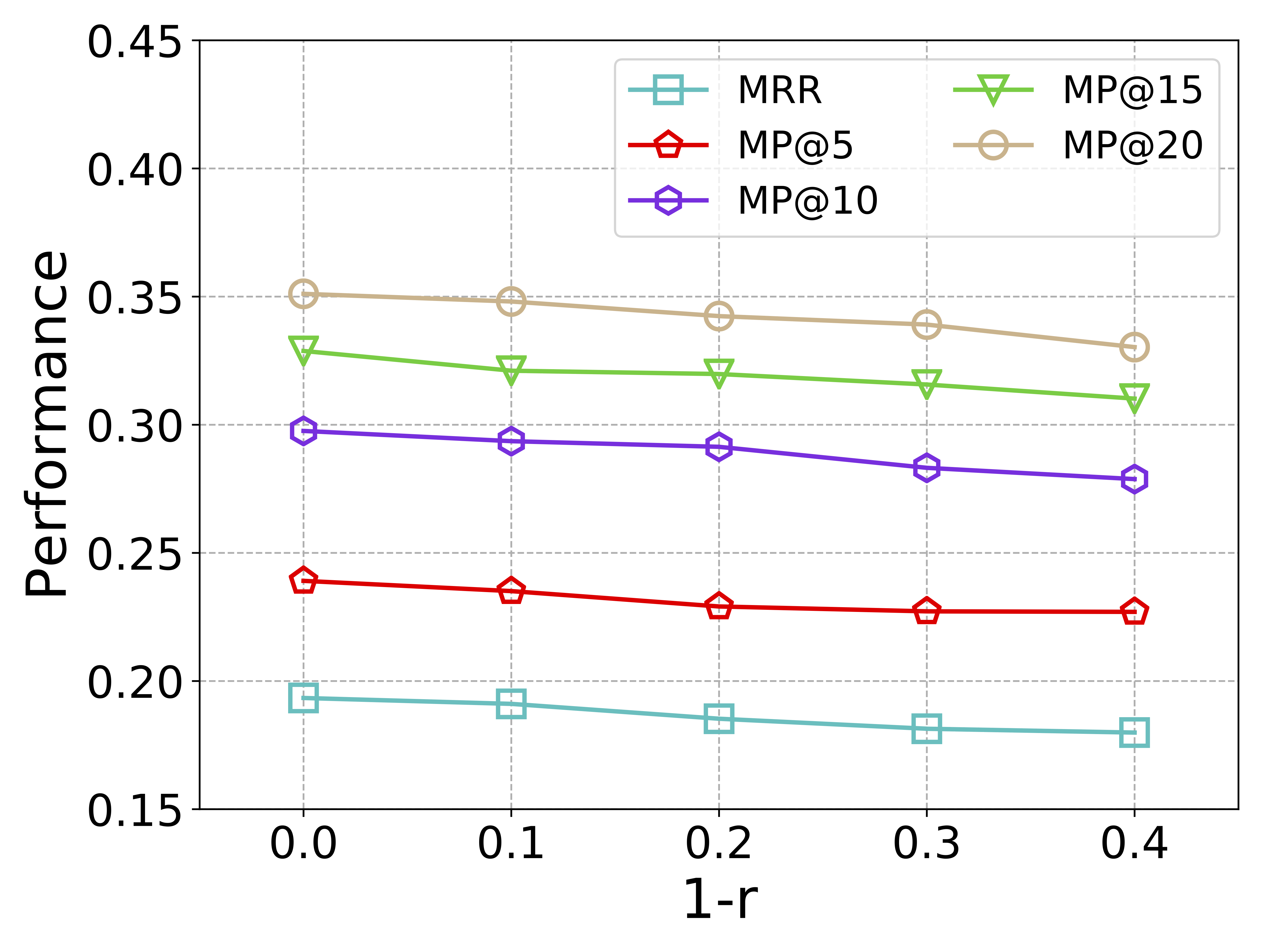}}  
\vspace{-2mm}
\caption{Robustness Comparison}\label{fig:robust}
\vspace{-6mm}
\end{figure}

In this subsection, we explore the robustness of different models against the sparsity of job transition graph. Specifically, we make the original graph sparser by subsampling the training edges at different rates $r=\{0.9,0.8,0.7,0.6\}$(only keep 90\%, 80\%, 70\%, 60\% edges). Then we retrain our model and baseline models on the subsampled graph and compare the link prediction performance degradation. Here we use IT dataset as an example. We compare our model Job2Vec with DeepWalk for conciseness. Job2Vec take four views (graph, semantic, transition balance and transition duration view) into account while DeepWalk only considers graph view. From Figure \ref{fig:robust}, we can observe that the performance of DeepWalk degrades sharply as $r$ decreases while Job2Vec seems to hold steady. This again confirms the effectiveness of incorporating more views especially semantic view over the sparsity issue of job transition graph. This can also be well explained: when the graph is sparse, many nodes in graph have poor connectivity, existing graph embedding models can not learn sufficient representations from graph view. But in semantic view, shared key words can connect different job titles in Job-Graph even though when it is sparse. Learning feasible representations from semantic view does not rely on graph connectivity and thus can perform excellent performance over sparse graph.

\begin{table}[t]
\abovecaptionskip
\belowcaptionskip
\caption{Job Title Benchmarking Cases.}
\small
\centering
\setlength\tabcolsep{2pt}
\scalebox{0.92}{
\begin{tabularx}{\linewidth}{|M{4.16cm}|M{4.0cm}|}
\hline
\multicolumn{2}{| c |}{\textbf{IT}}  \\ 
\hline
Project Manager & Product Manager PC Accessories \\
-IBM            & - Microsoft  \\
\hline
IT Support Lead and System Trainer & IT Support Lead  \\
 - HP                              &  - IBM \\
\hline
SWE - Google & Machine Learning Engineer - Airbnb \\
\hline
Software Engineer & Data Scientist\\
- Facebook            & - Microsoft \\
\hline
\multicolumn{2}{|c|}{\textbf{Finance}} \\
\hline
Investment Banking Analyst & Investment Banking Analyst\\
- Citi                     & - J.P. Morgan \\
\hline
Equity Research Analyst & Equity Research Analyst \\
- Nomura             & - Goldman Sachs\\
\hline
Financial Analyst & Financial Analyst \\
- Goldman Sachs   & - Rushmark Properties \\
\hline
Portfolio Manager & Portfolio Manager \\
- WellsFargo      & - ReMark Capital Group\\
\hline
\end{tabularx}
}
\label{tab:case}
\vspace{-0.5cm}
\end{table}

\subsection{Visualization}

\begin{figure}[t]
	\centering
	\subfigure[Deepwalk]{\label{fig:deepwalk}\includegraphics[width=0.45\linewidth]{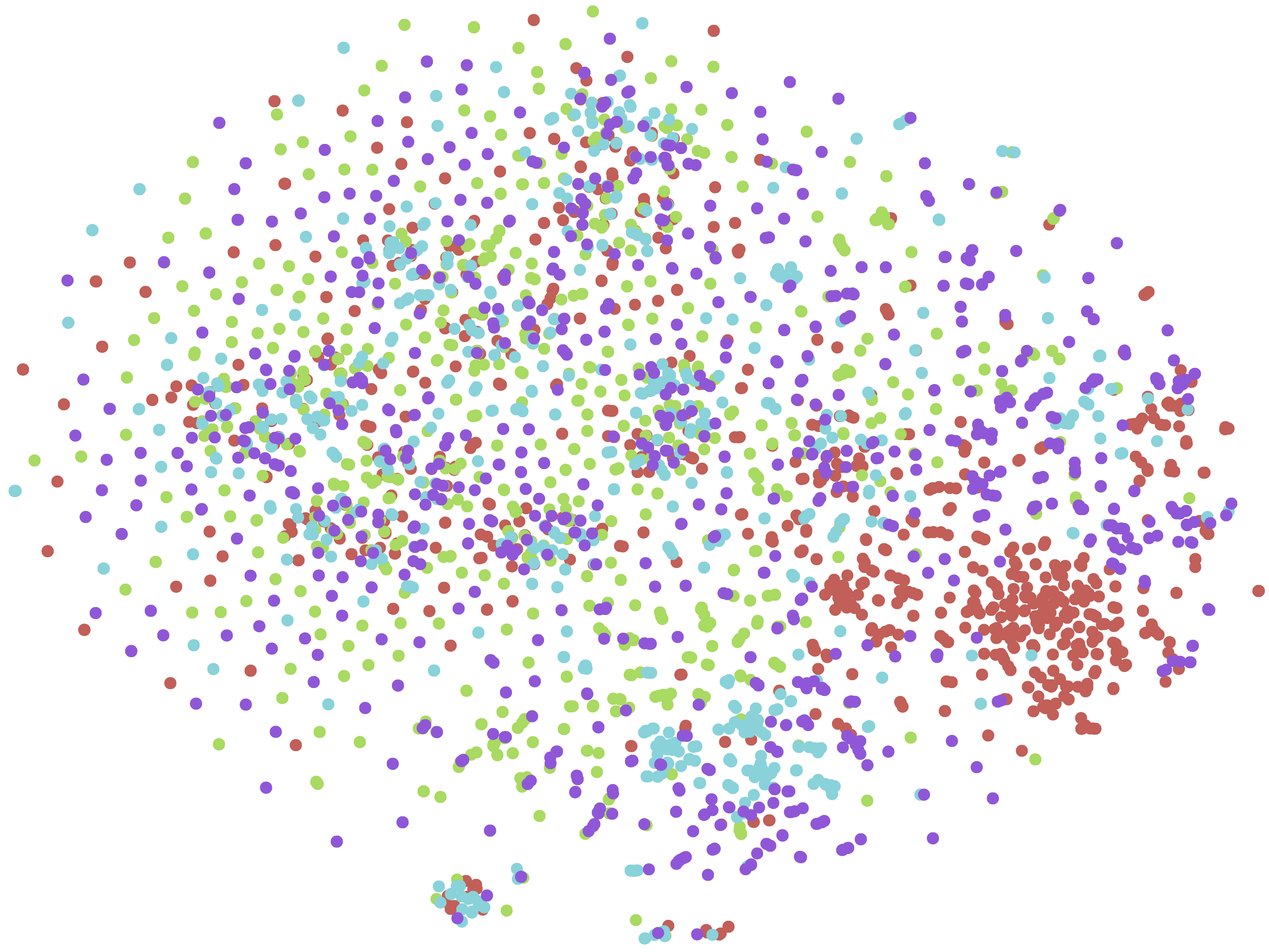}}
	\hspace{0.5cm}
	\subfigure[word2ec]{\label{fig:overall_precision_tokyo}\includegraphics[width=0.45\linewidth]{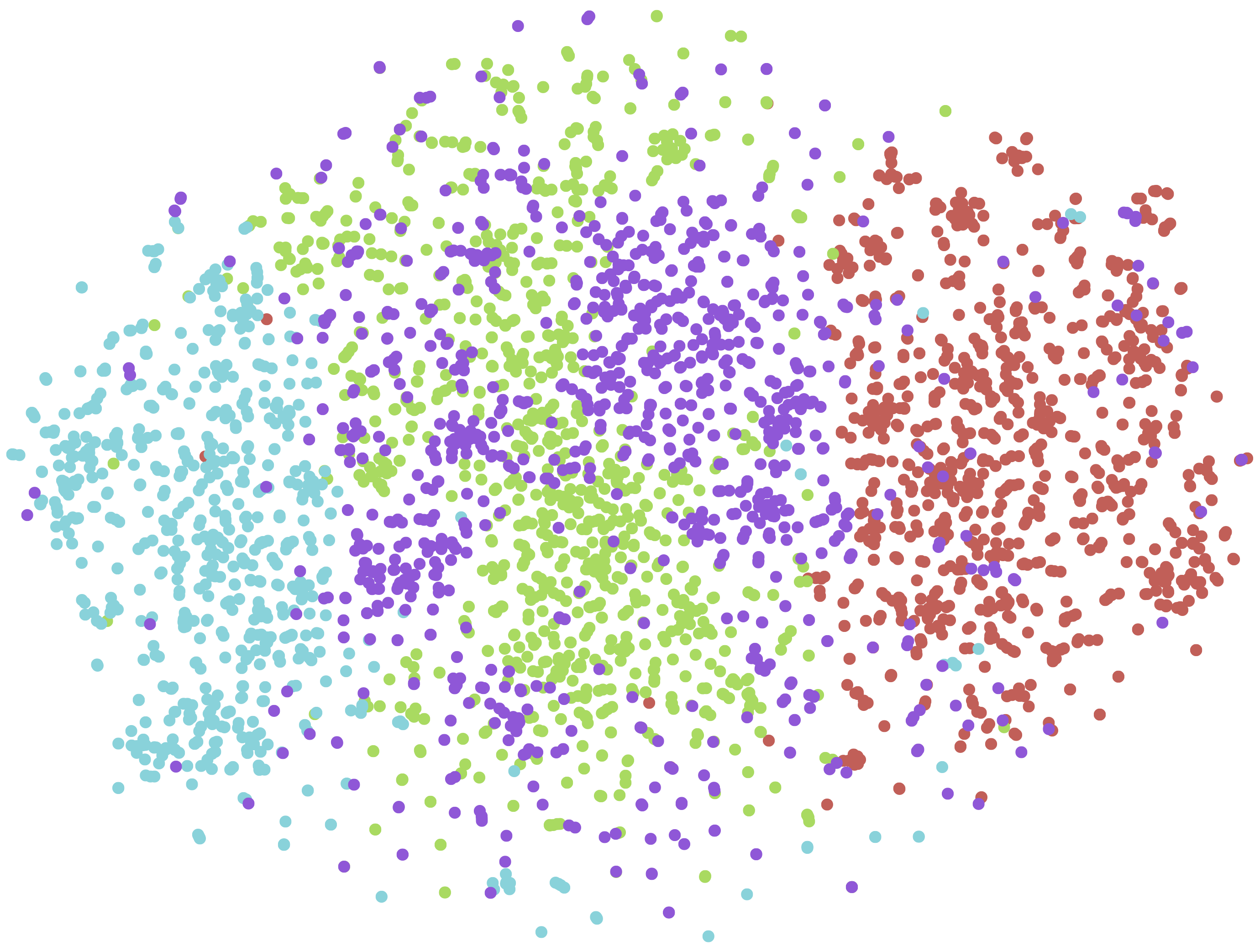}}
	\subfigure[LINE-1st]{\label{fig:LINE-1st}\includegraphics[width=0.45\linewidth]{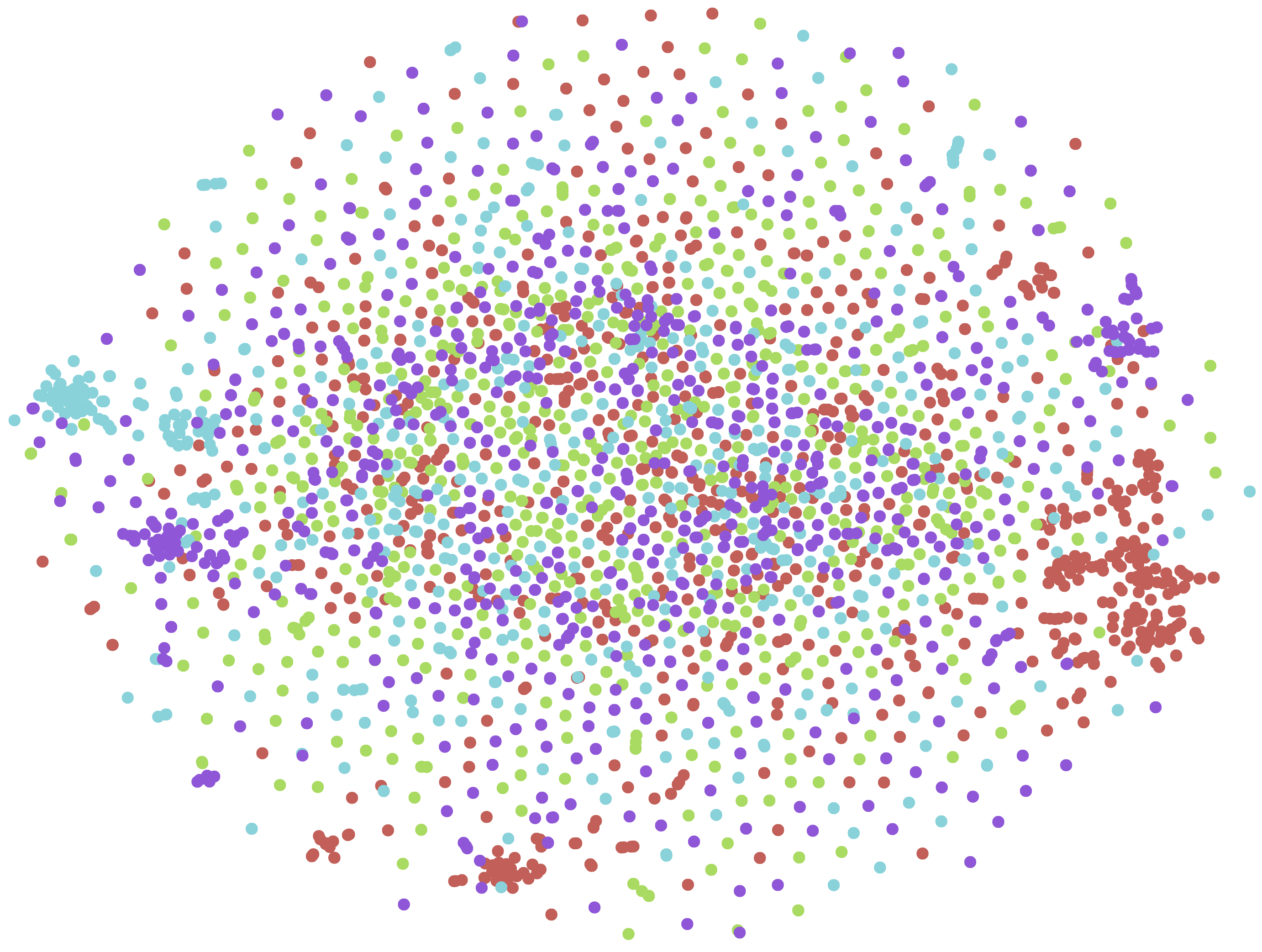}}
	\hspace{0.5cm}
	\subfigure[LINE-1+2]{\label{fig:LINE-1+2}\includegraphics[width=0.45\linewidth]{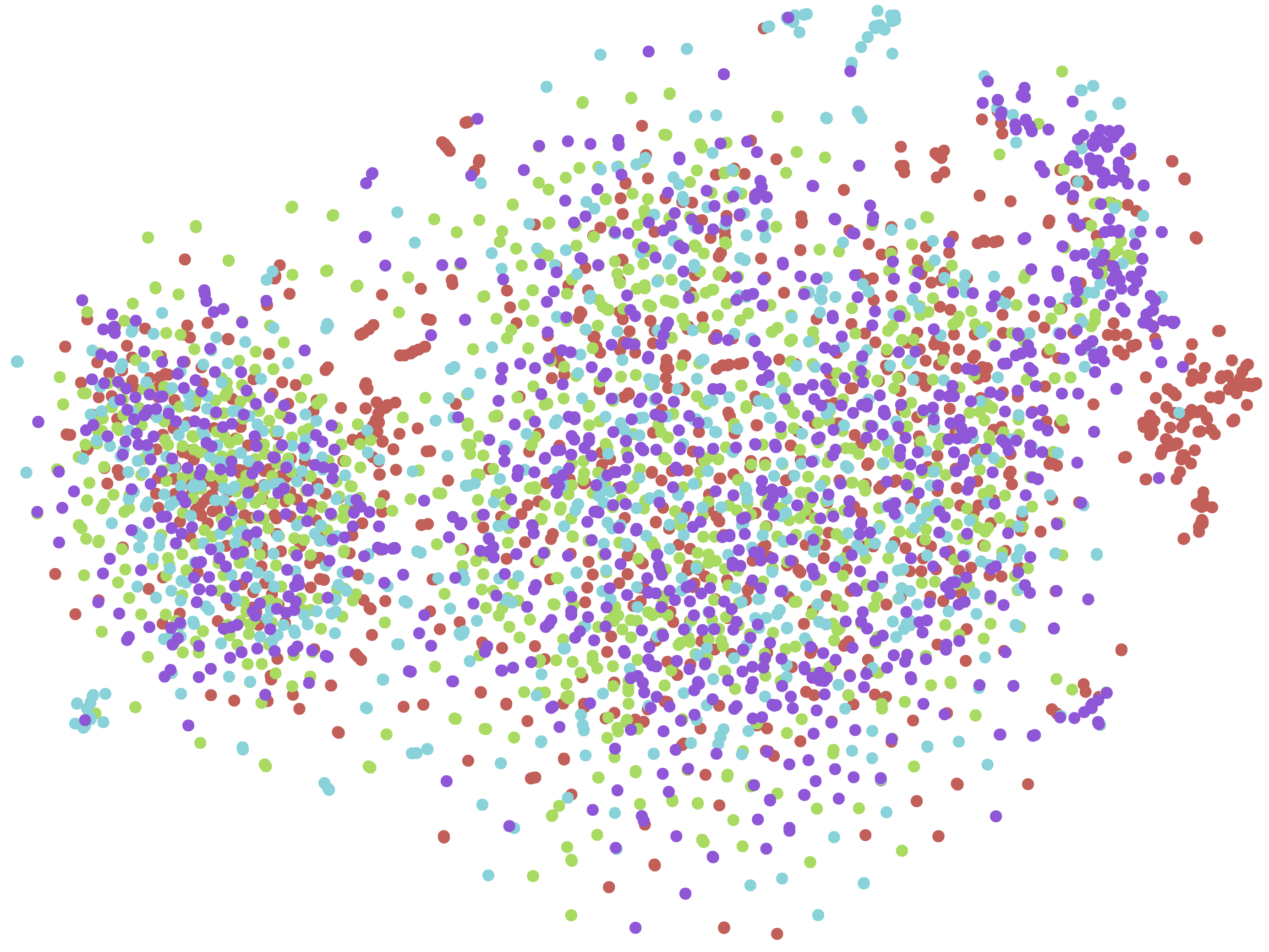}}
	\subfigure[node2vec]{\label{fig:node2vec}\includegraphics[width=0.45\linewidth]{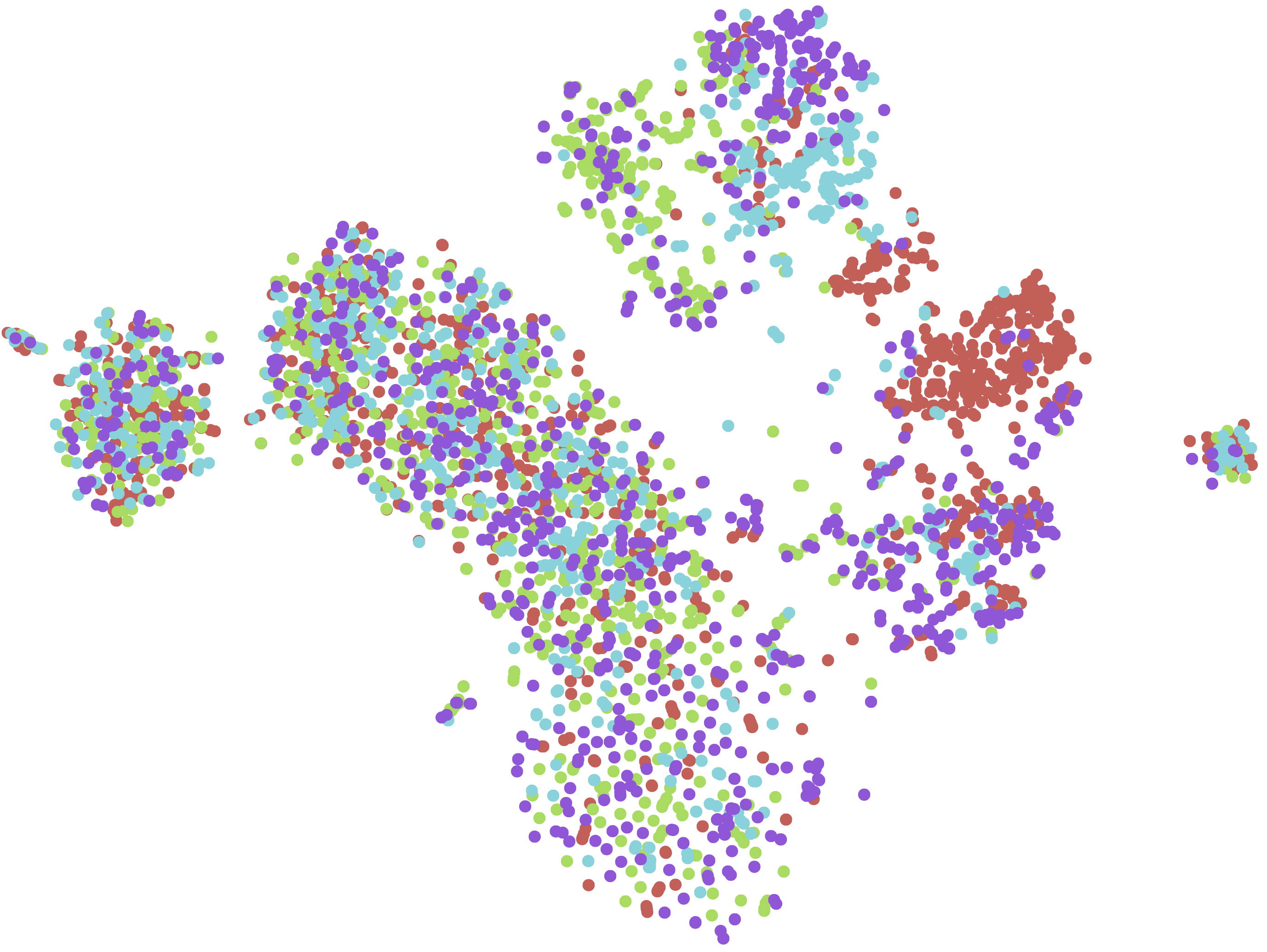}}
	\hspace{0.5cm}
    \subfigure[Job2Vec (Ours)]{\label{fig:job2vec}\includegraphics[width=0.45\linewidth]{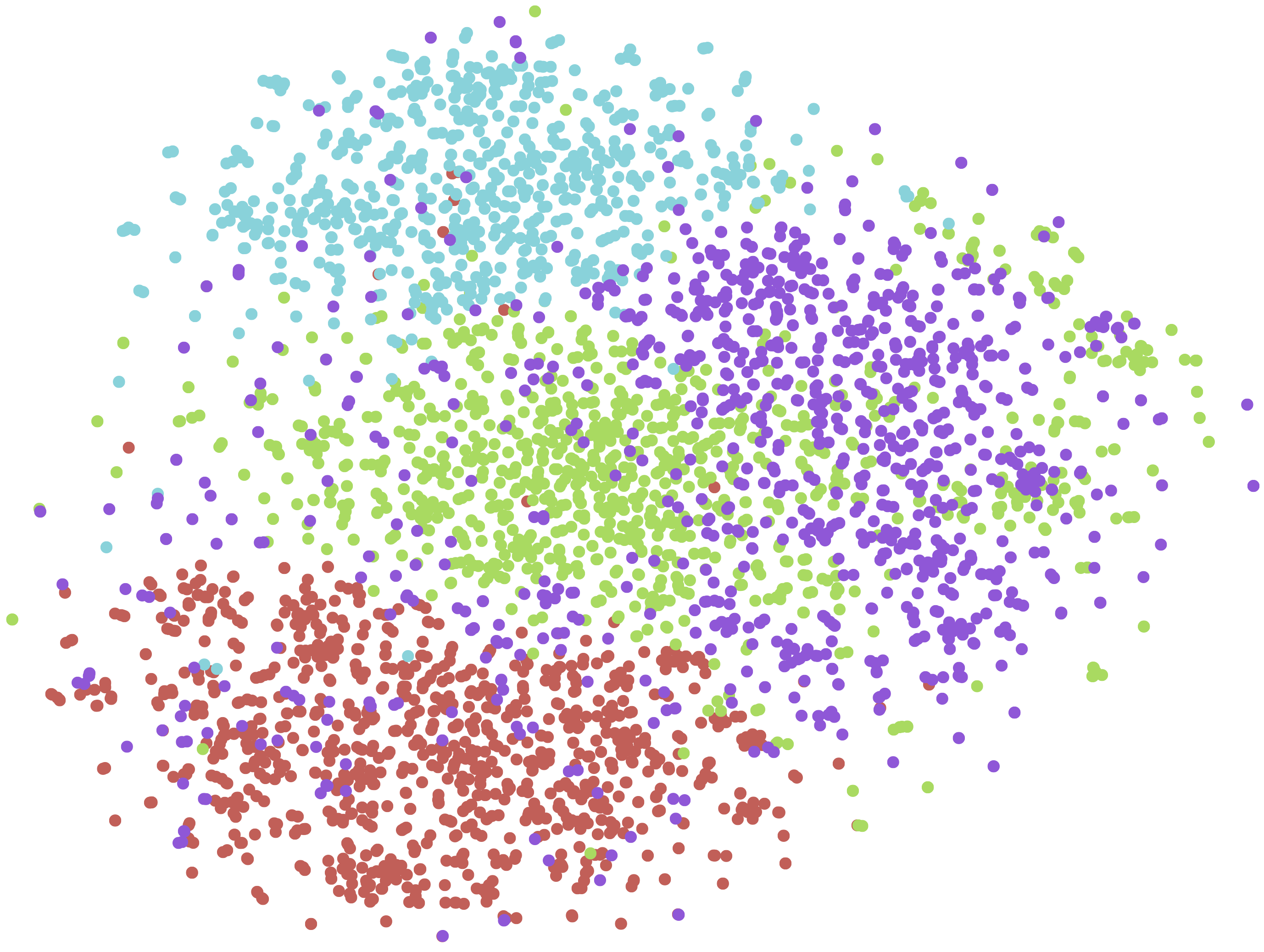}}
		\vspace{-0.25cm}
	\captionsetup{justification=centering}
	\caption{Visualization of the learned representations}
		\vspace{-0.35cm}
	\label{fig:visualization}
\end{figure}

We visualize the learned representations in Figure~\ref{fig:visualization} to show the promising benchmarking results of our proposed model.
For convenience, we select four categories of job title representations, including engineer, sales, consultant and manager.
We randomly sampled 1000 job titles for each categories.
We utilize t-SNE~\cite{van2014accelerating} to reduce the representation dimensions to do the visualization. Each color corresponds to one category of job titles.

Figure~\ref{fig:visualization} shows that our proposed Job2Vec achieves the best results. Each category of representations learned by our model can be clustered into four groups very well.
In another word, job titles are benchmarked by our proposed model effectively.
However, the representations learned by baselines are distributed randomly in the space which cannot reveal the becnmarking relations among job titles.
The potential explanation is that our proposed method jointly model four views to preserve the topology structure, semantics and job transition patterns, which plays an essential role in the task of job title benchmarking.


\subsection{Job Title Benchmarking Case Studies}
In this subsection, we show some Job Title Benchmarking (JTB) results extracted from existing Job Transition Graph as well as some link prediction JTB results generated by our model and baseline models.



\begin{table*}[t]
\abovecaptionskip
\belowcaptionskip
\caption{Top 3 Results of Link Prediction Comparison.}
\label{tab:top3}
\begin{center}
\small
\scalebox{0.92}{
    \begin{tabular}{  |c | c |c|}
    \hline
       & \multicolumn{2}{c|}{Top 3 results}  \\ \hline
     \multirow{3}{*}{Software Engineer-Facebook} & Job2Vec &SDE-Microsoft, Software Development Engineer-Amazon, Software Developer-Google  \\
    \cline{2-3}
     & DeepWalk     & Software Developer-Google, \textbf{Product Manager-Microsoft}, Software Test Engineer-Google
  \\\cline{2-3}
     & Word2Vec   &\textbf{Senior Software Engineer-Facebook}, Software Engineer-IBM, Software Engineer-Apple
   \\\hline
  \multirow{3}{*}{Project Manager-IBM} & Job2Vec & Product Manager-Microsoft, Project Manager-Microsoft, Program Manager-IBM\\\cline{2-3}
     & DeepWalk     &  Product Manager-Microsoft, \textbf{Senior Consultant-IBM}, \textbf{Storage Administrator-IBM}  \\\cline{2-3}
     & Word2Vec    &\textbf{Project Manager Lead-IBM}, Advisory Project Manager-IBM, Project Manager-Microsoft   \\\hline

    \end{tabular}
}
\normalsize
\end{center}
\vspace{-0.4cm}
\end{table*}

Table \ref{tab:case} shows eight JTB cases extracted from the aggregated Job-Graph, apparently job titles that have similar responsibilities and expertise level while also from companies of the same volume are matched. It is interesting that JTB can find some matching pairs that are not similar literally, such as (Software Engineer, Data Scientist), (SWE, Machine Learning Engineer).
Table \ref{tab:top3} shows the Top 3 link prediction results of different models for "Software Engineer-Facebook". Titles in bold font are wrong predictions. It can be observed that DeepWalk tends to make predictions that have similar neighborhood structure in Job-Graph, while Word2Vec inclines to job titles that are semantically similar. With only graph topology view, DeepWalk may make anomalous predictions such as "Product Manager-Microsoft". With only semantic view, Word2Vec are likely to predict repetitive job titles and miss some interesting matching pairs that are not very similar literally, such as (Software Engineer-Facebook, SDE-Microsoft). Our model, Job2Vec, incorporating graph topology, semantic, transition balance and transition duration views, makes more reasonable predictions.

\section{Related Work}
In this section, we review two categories of literatures that are related to this paper, namely research on data mining for career trajectory analysis, and research on graph embedding.

\noindent\textbf{Data Mining for Career Trajectory Analysis.} With the rise of Online Professional graphs (OPNs), Career Trajectory Analysis have been proved useful in many human resource management (HRM) problems\cite{xu2018dynamic,li2017prospecting}. For example, \citeauthor{xu2016talent} build a organizational level job transition graph from OPN data and proposed a talent circle detection method to identify the right talent sources for recruitment\cite{xu2016talent}. For better assessing the expertise level or rank of a job title, \citeauthor{xu2018extracting} proposed a Gaussian Bayesian graph to extract the job title hierarchy of an organization from employees' career trajectory data\cite{xu2018extracting} . In \cite{li2017nemo}, a contextual LSTM model is proposed to accurately predict an employee's next career move. However, few existing works pay attention to the problem of Job Title Benchmarking(JTB), which has broad application prospect in human resource management. To the best of our knowledge, we are the first to extract JTB insights from job transition graph.

\noindent\textbf{Graph Representation learning} graph representation learning assigns nodes in a graph to low-dimensional representations and effectively preserves the graph structure. Recently, a significant amount of progresses have been made toward this emerging graph analysis paradigm\cite{song2009scalable,TSS_Lichen_AAAI18,Seg_Lichen_TIP18,grover2016node2vec,tang2015line,leskovec2010signed,ou2016asymmetric}. Inspired by the success of representation learning in natural language processing~\cite{mikolov2013distributed,mikolov2013efficient,le2014distributed}, DeepWalk\cite{perozzi2014deepwalk} is the first extension of Word2Vec to graph analysis. It uses random walks to sample paths from a graph and treat paths as "sentences" to train a SkipGram model to keep graph Proximity into learned embeddings. Node2Vec~\cite{grover2016node2vec} generalizes DeepWalk by designing a more flexible random walk strategy. LINE\cite{tang2015line} proposes first-order and second-order proximity to keep graph properties and combines both by concatenating first-order and second-order vectors. To better model the asymmetric property of graphs, \citeauthor{ou2016asymmetric} proposes asymmetric proximity preserving (APP) graph embedding~\cite{ou2016asymmetric}. 
However, in our problem setting, we need to jointly model multi-view representations, and obtain an unified representation fused from multi-view.
Current graph representation learning cannot be directly applied into the JTB scenario.
To the best of our knowledge, our work is the first attempt to solve the JTB problem via multi-view graph representation learning.

\noindent\textbf{Multi-View Representation learning} has become attractive and urgent as the increasing multi-modal sensors are widely deployed in a great number of real-world applications~\cite{multiview_survey}. It explores the complementary information among different views, where the views refer to various feature representations, modalities or sensors. Most of the approaches focus on the multi-view data including features~\cite{AMGL_AAAI16}, images~\cite{MLAN_AAAI17}, and videos~\cite{EVaction_lichen,VCDN_lichen1}, while our approach reveal the information from multiple graph structure data which is challenging.

\vspace{-2mm}
\section{Conclusions}

In this paper, we propose a data-driven solution for the problem of job title benchmarking (JTB). We construct the job title transition graphs (Job-Graph) to represent job transitions, and reformulate the JTB problem as the task of link prediction over the Job-Graph.
Specifically, we propose a collective multi-view representation learning method by jointly learning the four views of representations, including graph topology view, semantic view, job transition balance view, and job transition duration view.
Besides, we also propose a encode-decode based fusion method to obtain an unified representation from the multi-view representations.
We present intensive experimental results with IT and finance related job transition data to demonstrate the effectiveness of our method.

\bibliographystyle{ACM-Reference-Format}
\bibliography{reference}
\end{document}